\documentclass{article}
\usepackage{spconf,amsmath,graphicx,hyperref}
\usepackage{booktabs}
\usepackage{subcaption}
\usepackage{caption}
\usepackage{xcolor}
\usepackage{float}
\usepackage{algorithm}
\usepackage{algorithmic}
\usepackage{multirow}
\usepackage{array}   
\usepackage{longtable}
\usepackage{bm}
\usepackage{acronym}
\usepackage{tikz}
\usepackage[table]{xcolor}
\ninept

\newcommand{\circled}[1]{\tikz[baseline=(char.base)]{
  \node[shape=circle,draw,inner sep=0.7pt] (char) {#1};}}

\title{Progressive Multi-Ancestor Bit-Depth Distillation }
\name{
Adil Mubashir Chaudhry$^{1}$,
Osama Ahmad$^{2}$,
Zubair Khalid$^{1}$,
Murtaza Taj$^{1}$
}

\address{
\small{$^{1}$LUMS School of Science and Engineering, $^{2}$University of Massachusetts Amherst, USA}\\
}
\begin{document}
%
\maketitle
%


\begin{abstract}

Model compression strategies are widely employed to reduce memory footprint and network complexity, particularly for devices with constrained computational, memory, and energy resources. Prior works that rely on simultaneous conversion from floating-point high-precision (FP32) to integer low-precision (INT4) representations and distillation into smaller models suffer from unstable training and drastic degradation of prediction performance. To address these limitations, we propose a unified framework, known as \textbf{P}rogressive \textbf{M}ulti-\textbf{A}ncestor \textbf{B}it-depth \textbf{D}istillation (PMABD), that progressively compresses the network while transferring knowledge through a growing pool of higher-precision ancestor teachers. PMABD generates a sequence of intermediate teachers that each learn from all higher-precision ancestors and jointly supervise the final target student. This multi-ancestor, multi-stage design  stabilizes ultra-low-bit quantization by lowering quantization noise profiles across training and ensuring stable quantization.
Experiments on CIFAR-10/100 with ResNet-20/32/18, and Tiny-ImageNet with MobileNetV2 show that PMABD outperforms state-of-the-art compression frameworks, results in 1.06$\%$ increase in performance of W2A2 (ResNet-18/CIFAR-100) student model. We show that a saturation-based stopping criterion contributes to improve the performance of our final student.

\end{abstract}


\section{Introduction}
\label{sec:intro}

Deep Neural Networks (DNNs) have demonstrated remarkable performance for various real-world applications, including image classification, object detection, and speech recognition~\cite{lecun2015deep}. However, with ever-increasing model depth and complexity, deploying deep learning models has been a significant challenge, particularly for resource-constrained edge devices~\cite{Han2016_DeepCompression,Jacob2018_IntegerQuant}. This motivates us to explore model compression approaches so that DNNs can execute on edge devices with minimal accuracy degradation.


Recent works have explored techniques to obtain low-bit target student networks through knowledge distillation~\cite{pham2023collaborative,Zhu2023QFD}. Frameworks that reduce precision directly from FP32 to INT4 drastically degrade overall performance and introduce substantial quantization error. Although progressive quantization (FP32$\rightarrow$INT8$\rightarrow$INT4) shows improved results over direct reduction~\cite{zhuang2021effective,zhuang2018towards}, the model's performance monotonically decays, and it is difficult to recover accuracy at subsequent lower precision levels. Multiple and intermediate teachers have been extensively used to provide flexible and appropriate guidance for improved knowledge distillation~\cite{mirzadeh2020improved, you2017learning, liu2020adaptive, haase2025hpm}.
\begin{figure}[t]
    \centering
\includegraphics[width=0.98\linewidth]{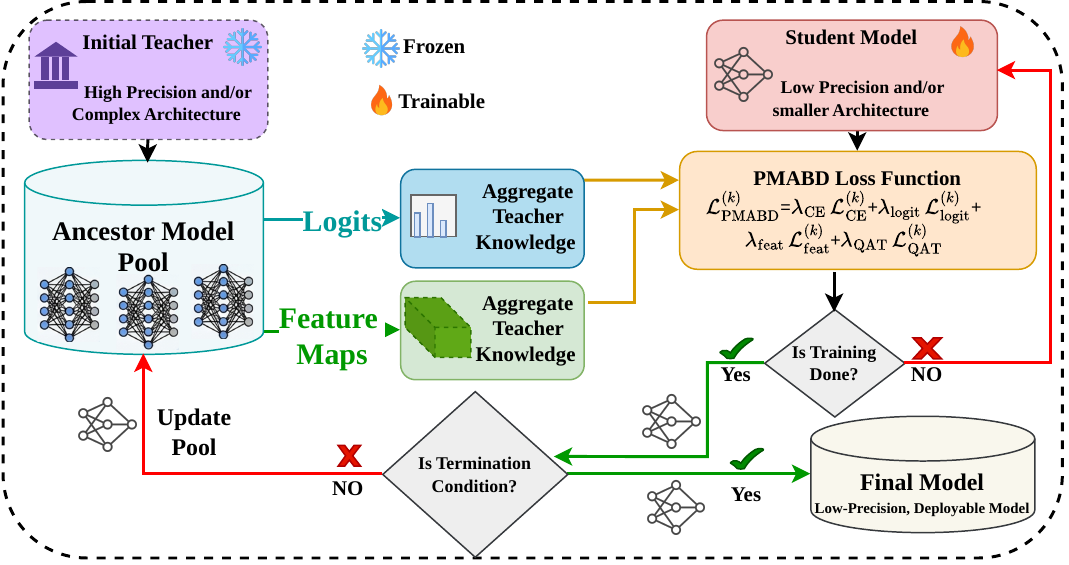}
    \caption{PMABD block diagram illustrating the progressive loop. A shared model pool aggregates logits and feature-map knowledge from multiple ancestor teachers to guide the student model, PMABD loss function $\mathcal{L_{\rm{PMABD}}}$ leads the training phase of the model. \textit{Is Training Done?} signifies that each model is trained until early stop criteria is achieved.  \textit{Is Termination condition?} provides deployable final model if our desired precision model is acquired and accuracy is saturated between two consecutive models at same precision. Teachers are fake-quantized, so their logits and features are FP32 tensors; aggregation is done in FP32. }
    \label{fig:pmabd_block}
    \vspace{-1em}
\end{figure}

For model compression, recent advanced approaches use distillation to train low-precision student networks~\cite{boo2021stochastic, lee2025unifying, Mishra2018Apprentice}. Quantization-aware knowledge distillation (QKD) is a multi-phase training framework in which a fixed full-precision teacher distills knowledge to a low-bit quantized student~\cite{kim2019qkd}. QKD does not explicitly model the progressive reduction of precision levels, which can lead to unstable optimization and increased quantization noise. CMT-KD aggregates knowledge from multiple quantized teachers through learned importance weights~\cite{pham2023collaborative}. It treats teachers as independent knowledge sources and jointly optimizes multiple teachers and students within a single training stage. As a result, the knowledge transfer process does not explicitly model how representations evolve across precision levels, and substantial distribution shifts arise between high-precision and ultra-low-bit networks.

Divide-and-Conquer Quantization (DCQ) partitions teachers into several sections, each of which trains the student using its intermediate features; the sections are then combined to form a complete low-bit student model~\cite{elthakeb2020divide}. Quantized Feature Distillation (QFD) distills knowledge from the features of a quantized teacher by aligning the teacher network with the limited representational capacity of the student model, and demonstrates performance improvement over conventional knowledge distillation~\cite{Zhu2023QFD}.
SQAKD is a framework that reformulates QAT as a co-optimization problem that jointly minimizes the Kullback-Leibler (KL) divergence between full-precision and low-bit model outputs and the discretization error, without supervision from labels~\cite{zhao2024self}. Unified Progressive Quantization (UPQ) introduces a sequential quantization pipeline in which an intermediate INT4 PTQ model serves as a warm-start initialization for subsequent INT2 quantization~\cite{lee2025unifying}. However, the intermediate INT4 model is used only for initialization, while knowledge distillation is performed solely from the single original FP16 teacher to the INT2 student.
The primary contributions of this work are as follows:
\begin{itemize}
\vspace{-0.5em}
\item We propose a novel multi-stage distillation framework that progressively lowers the precision of high-precision models by accumulating the knowledge of all ancestor models. Leveraging ancestor knowledge allows the framework to recover information lost during distillation and achieve better training stability throughout the quantization and yield a smoother optimization landscape for the student model.    

    \item After a complete training run, the framework produces multiple well-performing models at different precision levels, available for deployment across devices with varied compute and memory requirements. 
\vspace{-0.5em}
    \item  Extensive empirical evaluation comparing the PMABD framework against standard optimization techniques on ResNet-32/20/18, and MobileNetV2. PMABD is an architecture agnostic framework; it is compatible with any CNN architecture and supports cross-architecture training at any precision stage.  Results demonstrate that progressive multi-ancestor supervision yields higher accuracy and stability.


\end{itemize}

\section{Method}
\label{sec:method}
\textit{Progressive Multi-Ancestor Bit-Depth Distillation (PMABD)} framework is formulated as a \textbf{progressive loop} that generates lower-precision student models by aggregating knowledge of the variable-precision teachers from the model pool, as illustrated in Fig.~\ref{fig:pmabd_block}.

Given a frozen teacher $T_1$ at certain high precision $b_1$, our objective is to achieve a $T_k$ model having precision $b_k$ ($b_k < b_i \quad \forall\, i < k$). Here, all ancestor teachers $\{T_1,T_2,\dots,T_{k-1}\}$ from the model pool distill their knowledge to $T_k$ model. We can consider two successive teachers can have the same precision $b_1\geq b_2 \geq \dots \geq b_k$ albeit $b_1 > b_k$ its hold for our desired output model. The models at specific precision $b_i$ are denoted by $\mathcal{M}_i$.

\subsection{Model Architecture}
PMABD employs a three distinct model level, where each model level fulfills a distinct role in the knowledge distillation and precision reduction process:

\textbf{\circled{1} Teacher Models  -} PMABD requires at least one high-capacity frozen FP32 teacher ($\mathcal{M}_1$) to anchor the distillation chain, which provides high representational capacity and serves as a stable semantic anchor throughout all subsequent stages.

\textbf{\circled{2} Intermediate Models -} Intermediate models serve the dual purpose of being trained students and, once training is complete, contribute as a teacher in distillation chain. The architecture of the intermediate models can be chosen independently from the teacher(s) or the final student.

\textbf{\circled{3} Student Model -} The final student is the target low-bit ($b_k$) model to be deployed. In our proposed framework its architecture is independent, no architectural constraint, of teacher and intermediate models. 

The framework begins with a pretrained FP32 teacher model added to an initially empty pool ($\mathcal{P}$).  The loop proceeds by selecting a target bit-depth lower than any existing model in the pool, training a new student at that bit-depth under supervision from all current pool members, and then inserting the trained student into the pool $\mathcal{P} \leftarrow \mathcal{P} \cup \{T_k\}$. This process repeats until the target precision is reached. Each teacher distills its logits and features to successive student model until patience-based early stopping criteria is achieved.  The specific precision ladder, number of stages, and architecture at each stage are configuration choices, not structural requirements of the framework. 
However, the criteria to identify our required student model at a desired precision is decided based on the saturation condition, which is verified using accuracy ($T_k$) $\leq$ accuracy ($T_{k-1}$), if it is fulfilled, we treat $T_{k-1}$ as our desired student model. A tie means the new stage brings no gain, so we stop and keep the earlier, cheaper model.
\subsection{Loss Formulation and Training Objective}
The loss function $\mathcal{L}_{\rm{PMABD}}^{(k)}$ for current student model is employed to train PMABD framework,
consisting of four components
\begin{multline}
\label{eq:pmabd_loss}
\mathcal{L}_{\rm{PMABD}}^{(k)}
= \lambda_{\rm{CE}}\,\mathcal{L}_{\rm{CE}}^{(k)}
+ \lambda_{\rm{logit}}\,\mathcal{L}_{\rm{logit}}^{(k)} \\
+ \lambda_{\rm{feat}}\mathcal{L}_{\rm{feat}}^{(k)}
+ \lambda_{\rm{QAT}}\, \mathcal{L}_{\rm{QAT}}^{(k)},
\end{multline}
where $\lambda_{\rm{CE}}$, $\lambda_{\rm{logit}}$,  $\lambda_{\rm{feat}}$, and $\lambda_{\rm{QAT}}$  are the penalty terms for cross-entropy loss $\mathcal{L}_{\rm{CE}}$, logit distillation, feature distillation, and QAT loss, respectively.
 Here, logit loss $\mathcal{L}_{\rm{logit}}^{(k)}$ is formulated in terms of Kullback-Leibler (KL)-divergence is given as:
\begin{equation}
\mathcal{L}_{\mathrm{logit}}^{(k)}
= \mathrm{KL}\!\left(
 \big( \sum_{i=1}^{k-1} \gamma_i\,  p_{T_i}^{(\tau)} \big) \,\big\|\, p_{T_k}^{(\tau)}
\right) \tau^2,
\end{equation}
where $\tau$ is the distillation temperature, $p_{T_i}^{(\tau)}$ is the softened distribution and $\gamma_i$ is the dynamically weighting of each teacher $i$.
Let $f_l(T_i)$ denote the feature maps of the $i^{\rm{th}}$ teacher  at layer $l$. The feature alignment loss is defined as
$\mathcal{L}_{\rm{feat}}^{(k)} = \mathcal{L}_{\rm{RKD}}^{(k)}~\cite{park2019relational} + \mathcal{L}_{\rm{AT}}^{(k)}~\cite{zagoruyko2017payingattentionattentionimproving},$\footnote{
\begin{align*}
    \mathcal{L}_{\rm{RKD}}^{(k)} = \frac{1}{L}\sum_{l}\Bigl[
  \ell_{\rm{H}}\!\left(\tilde{d}^s_l,\, \tilde{d}^t_l\right)
+ 2\,\ell_{\rm{H}}\!\left(\cos\Theta^s_l,\, \cos\Theta^t_l\right)
\Bigr], \\
\mathcal{L}_{\rm{AT}}^{(k)} = \frac{1}{L}\sum_{l}
\left\|
\frac{Q^s_l}{\|Q^s_l\|_2} - \frac{Q^t_l}{\|Q^t_l\|_2}
\right\|_2^{2},
\end{align*}
where $\tilde{d}$ denotes mean-normalised pairwise distances between sample embeddings, $\cos\Theta$ denotes pairwise cosine similarities, and $\ell_{\rm{H}}$ is the Smooth-$\ell_1$ (Huber) loss. $Q_l^t = \mathrm{vec}(\bar{f}_l)$  is the averaged teacher attention map in vector form and $Q_l^s =\mathrm{vec}(f_l)$ is the attention map of current student.}  
where both terms compare the projected student features $\phi(f_l(T_k))$ against the mean teacher feature $\bar{f}_l = \frac{1}{k-1}\sum_{i=1}^{k-1} f_l(T_i)$ at each layer $l$, and $\phi(\cdot)$ is a $1{\times}1$ convolution followed by Batch Normalization that aligns the student's channel dimensions to the teacher's. The quantization regularization term penalizes excessive clipping of weights and activations and is defined as:
\begin{equation}
\mathcal{L}_{\rm{QAT}}^{(k)}
= \sum_{l}\Bigl(
  \|\hat{W}_l - W_l\|_2^2 + \|\hat{A}_l - A_l\|_2^2
\Bigr),
\end{equation}
where $W_l$, $A_l$ are the full-precision weights and activations at layer $l$, and $\hat{W}_l$, $\hat{A}_l$ are their fake-quantized counterparts. This penalises the mean-squared quantization error directly. 

\section{Experiments}
\vspace{-0.3cm}
\subsection{Experimental Setup \& Datasets}
\label{sec:setup}
\vspace{-0.2cm}
All experiments are designed using PyTorch and TorchVision on RTX3070 Ti (8GB)  GPU hardware.
Benchmarking datasets are Tiny-ImageNet and CIFAR-10/100. 
All backbone teacher models are trained from scratch on the train split. For the baseline FP32 variants, we 
use the same SGD recipe from the CIFAR model hub\cite{chenyaofo_pytorch_cifar_models}: 
SGD with Nesterov momentum (learning rate $\text{lr}=0.1$, momentum~$=0.9$,
weight decay~$=5\times10^{-4}$), CosineAnnealingLR ($E_\text{max}=200$,
lr$_\text{min}=0$), plain cross-entropy, while at other stages we employ AdamW optimizer.  The batch size is 32. On Tiny-ImageNet, PMABD ladder runs without saturation.  
Warmup, to improve early training stability and reduce the risk of unstable parameter updates, is 2 epochs for FP32 distillation, 3 for W8A8 and W8A4, 5 for W4A4, 4 for W4A3, 5 for W3A3, 6 for W3A2 and 8 for W2A2. The distillation temperature is 4 for FP32 and W8A8, 5 for W8A4 to W4A3, 5.5 for W3A3, and 6 for W3A2 and W2A2. 
Each reimplemented baseline starts from the same FP32 teacher checkpoint as PMABD and is trained for the same number of epochs per precision stage, so accuracy differences reflect the distillation method rather than the teacher or the training budget. CIFAR-100 train data (50k) split into 45K (\texttt{train\_eff}), validation set 5k.   
The code is available at \underline{\href{https://github.com/adil-mubashir-ch/PMABD}{https://github.com/adil-mubashir-ch/PMABD}}. 
\begin{table}[t]
\centering
\caption{PMABD precision ladder for ResNet-32 / CIFAR-100. FP32 models are trained from scratch on the 45k split with the SGD recipe of Sec.~\ref{sec:setup}; quantized stages use LSQ+~\cite{bhalgat2020lsq+} fake quantizers and AdamW. $^\ast$Converged student of level $\mathcal{M}_3$.}
\label{tab:pipeline}
\resizebox{0.9\columnwidth}{!}{
\begin{tabular}{lllll}
\toprule
\textbf{Level} & \textbf{Model} & \textbf{Arch.} & \textbf{Precision} & \textbf{Init.\ / Teachers} \\
\midrule
$\mathcal{M}_1$ & $T_1$ & ResNet-56 & FP32 & Scratch / none (frozen) \\
$\mathcal{M}_1$ & $T_2$ & ResNet-32 & FP32 & Scratch / none \\
$\mathcal{M}_2$ & $T_3$ & ResNet-32 & W8A8 & $T_2$ / $T_1, T_2$ \\
$\mathcal{M}_3$ & $T_4$ & ResNet-32 & W8A4 & $T_3$ / $T_1$--$T_3$ \\
$\mathcal{M}_4$ & $T_5$ & ResNet-32 & W4A4 & $T_4^\ast$ / $T_1$--$T_4$ \\
$\mathcal{M}_5$ & $T_6$ & ResNet-32 & W2A3 & $T_5$ / $T_1$--$T_5$ \\
$\mathcal{M}_6$ & $T_7$ & ResNet-32 & W2A2 & $T_6$ / $T_1$--$T_6$ \\
\bottomrule
\end{tabular}
}
\vspace{-1em}
\end{table}
\begin{table}[t]
\centering
\caption{W4A4 comparison with published results. All baseline numbers are taken directly from the cited papers. 
$^\dagger$Our FP32 checkpoint, trained from scratch on our 45k split to avoid validation leakage. Sat.: saturation stopping; HP: high patience ($40$), LP: low patience ($20$).}
\label{tab:sota_w4a4}
\resizebox{0.85\columnwidth}{!}{
\begin{tabular}{lcc}
\toprule
\multirow{2}{*}{\textbf{Method}} & \textbf{ResNet-20 /} & \textbf{ResNet-32 / } \\
&CIFAR-10&CIFAR-100\\
\midrule
FP32 baseline (32/32) & 92.45 & 72.17$^\dagger$ \\
\midrule
Baseline Quantization~\cite{Zhu2023QFD} & 92.09 & 70.13 \\
Feature KD~\cite{Zhu2023QFD}            & 91.12 & 70.23 \\
EWGS~\cite{Lee2021EWGS}                 & 92.40 & 70.50 \\
Logit KD~\cite{Zhu2023QFD}              & 92.02 & 71.14 \\
APRT~\cite{Mishra2018Apprentice}       & 92.20 & 71.50 \\
QFD~\cite{Zhu2023QFD}                   & \textbf{93.07} & 71.52 \\
SQAKD (EWGS)~\cite{zhao2024self}        & 92.59 & 71.65 \\
\midrule
\rowcolor{gray!20}
\textbf{PMABD HP + Sat. (Ours)}              & --    & \textbf{71.87} \\
\rowcolor{gray!20}
\textbf{PMABD LP + Sat. (Ours)}               & --    & 71.62 \\
\rowcolor{gray!20}
\textbf{PMABD HP + No Sat. (Ours)}          & 92.53 & 71.68 \\
\bottomrule
\end{tabular}
}
\vspace{-2em}
\end{table}

\begin{table}[t]
\centering
\caption{Comparison of knowledge distillation strategies on ResNet-18 / CIFAR-100 under different quantization bit-widths. Accuracy for FP32 Baseline is 78.16. $E_{\max}$ is 50. ${^*}$ results reported in the original papers, ${^{**}}$ re-implemented under our training recipe.
}
\label{tab:ablation_kd}
\resizebox{.80\columnwidth}{!}{
\begin{tabular}{lcccc}
\toprule
\textbf{Method} & \textbf{W2A2} & \textbf{W3A3} & \textbf{W4A4} & \textbf{W8A8} \\ 
\midrule
Baseline${^*}$~\cite{Zhu2023QFD}              & 67.90 & 71.56 & 71.50 & -- \\
Feature KD${^*}$~\cite{Zhu2023QFD}            & 68.94 & 71.58 & 72.03 & -- \\
Logit KD${^*}$~\cite{Zhu2023QFD}              & 69.47 & 71.59 & 72.68 & -- \\
QFD${^*}$~\cite{Zhu2023QFD}                   & 70.63 & 72.45 & 73.16 & -- \\
LQ-Nets${^*}$~\cite{3491440.3491732}          & 70.80 & 72.90 & --    & -- \\
LQW+CAQ${^*}$~\cite{3491440.3491732}          & 72.10 & 73.00 & -- & -- \\
\midrule
CMT-KD$^{\dagger**}$~\cite{pham2023collaborative} & 75.76 & -- & 76.89 & 76.91 \\
DAQ-AKD$^{\dagger**}$~\cite{kur2025data}               & 76.12 & -- & 77.97 & 77.52 \\
SQAKD$^{\dagger**}$~\cite{zhao2024self}                  & 76.64 & -- & 77.64 & 77.41 \\
\midrule
\rowcolor{gray!20}
\textbf{PMABD (Ours)}$^{\dagger}$                   & \textbf{77.70} & \textbf{78.62} & \textbf{80.07} & \textbf{79.03} \\
\bottomrule
\multicolumn{5}{l}{\footnotesize $^\dagger$50 epochs per precision stage; FP32 teacher: 78.16\%.} \\
\vspace{-3.5em}
\end{tabular}
}
\end{table}
\subsection{Pipeline Stages}
\label{sec:pipeline}
\vspace{-0.2cm}
PMABD trains a chain of students at decreasing precision (Tab.~\ref{tab:pipeline}). Each student is initialized from the best checkpoint of the previous stage and distilled from every model in the pool. Only students at precision level $\mathcal{M}_3$ and $\mathcal{M}_4$ are trained under the saturation condition.

Transitioning from high precision model ($\mathcal{M}_1$) to low precision models $\{\mathcal{M}_2,\mathcal{M}_3,...\mathcal{M}_j\}$ is carried out using quantization aware training (QAT) method. To approximate the gradients during training, we adopt Learned Step Size Quantization (LSQ)~\cite{Esser2020LSQ}, where the quantization scale is a learnable parameter updated jointly with the model weights. 
Gradients through the rounding operator are approximated via straight through estimator (STE)~\cite{bengio2013estimating}, while the clipping boundary yields a zero gradient for out-of-range inputs.

In this implementation, the logit penalty $\lambda_{\rm{logit}}$ follows a cosine annealing schedule~\cite{Loshchilov2017_SGDR} with a lower-bounded floor. 
Concretely, at training epoch $t$ out of $E_{\max}$ total epochs:
\begin{equation}
\lambda_{\mathrm{logit}}(t) = \alpha\!\left(\delta + (1-\delta)\cdot\frac{1+\cos(\pi\, t/E_{\max})}{2}\right),
\vspace{-1em}
\end{equation}
where $\alpha$ is the per-stage base KD weight, and $\delta=0.15$ is a floor that keeps the teacher signal active throughout training.
Without this floor, the KD weight decays to zero in late epochs, consistently degrading validation accuracy as the student loses teacher regularization. Inspired by EDT~\cite{zhou2020channel}, the floor prevents this collapse; its value is a conservative heuristic rather than a tuned hyperparameter. The complementary $\lambda_{\rm{CE}}$ weight is $1 - \lambda_{\mathrm{logit}}(t)$. Two patience regimes are evaluated: high patience ($P=40$) and low patience ($P=20$). The random seed $42$ is used for weight initialization and data splitting. For the teacher Weighting scheme at every stage, we use entropy-based weighting.

\section{Results}
\label{sec:results}
\vspace{-0.3cm}

\subsection{Comparison with Baselines}
\label{sec:sota_resnet32}
\vspace{-0.2cm}
 
 As shown in Table~\ref{tab:sota_w4a4}, PMABD exceeds the SOTA by $+0.22\%$ absolute on the ResNet-32 / CIFAR-100 W4A4 benchmark. PMABD achieves $\mathbf{71.87\%}$, surpassing the state-of-the-art (SOTA) of $71.65\%$ from SQAKD \cite{zhao2024self}. PMABD outperforms other re-implemented SOTA frameworks because its progressive, multi-ancestor supervisory signal steer the student network into a wider and flatter optimization basin that is significantly more robust to quantization noise.However, at high patience and saturation improves the training of our desired student because at low learning rate high patience value leads training to better local minima.

\noindent\textbf{Saturation, data, and warm starts.} Saturation stopping lifts ResNet-32/CIFAR-100 W4A4 accuracy from 71.68\% to 71.87\% under high patience (Table~\ref{tab:sota_w4a4}), though we have only tested it in this one setting. PMABD also trains on just 45k CIFAR-100 images, holding out 5k for validation, while the published baselines use all 50k; it reaches competitive accuracy with less data. Finally, each precision level warm-starts from the one before it, so lower-precision models begin training already partly adapted to quantization noise rather than starting from FP32 weights.

We also represent the performance of PMABD on different precision, such as W2A2, W3A3, and W4A4 (Table~\ref{tab:ablation_kd}) on ResNet-18/ CIFAR-100. ResNet-18 contains $\sim$11.2M parameters, compared to the $\sim$0.27M and $\sim$0.46M parameters of ResNet-20 and ResNet-32, and this higher representational capacity helps prevent severe bottlenecking at lower precisions. It requires a total of 7 parents to train W2A2, including 3 mixed-precision models such as W8A4, W4A3, and W3A2. PMABD is trained with no saturation with entropy based policy. It shows significantly improved performance on W4A4; however, the accuracy drops at W3A3 from $80.07\%$ to $78.62\%$.

To study the effect of our proposed framework on a scalable dataset, we show the performance of PMABD on Tiny-ImageNet in Table~\ref{tab:tinyimagenet}. It indicates that the accuracy of PMABD on MobileNet-V2/W2A2 increases as compared to the baseline models.  Every method starts from the same FP32 checkpoint, but each PMABD precision starts from the previous precision level, giving it a warm start. PMABD also generates intermediate precision checkpoints for ultra-low bitwidths, gradually quantizing weights and activations separately. This indicates that all ancestor models contribute to reducing quantization error in our final student model (W2A2). In a training cycle, PMABD provides the best weights for four different quantization levels, whereas SQAKD takes four training cycles, with only $\approx1.5\times$ training time compared to SQAKD models.

\begin{table}[t]
\centering
\caption{Comparison of knowledge distillation strategies on MobileNet-V2/ Tiny-ImageNet~\cite{le2015tiny} under different quantization bit-widths. Baseline methods are implemented by us. The mean $E_{\max}$ for every precision stage is 50. Training hrs denotes the total number of GPU hours to obtain the W2A2 precision student. 
}
\label{tab:tinyimagenet}
\resizebox{.9\columnwidth}{!}{
\begin{tabular}{lcccc|c}
\toprule
\textbf{Method} & \textbf{W2A2} & \textbf{W3A3} & \textbf{W4A4} & \textbf{W8A8} & \textbf{Training hrs ($\downarrow$)}\\ 
\midrule

SQAKD\cite{zhao2024self}             & 2.61 & 60.10 &\textbf{69.96} & \textbf{75.38}& \textbf{16.57}\\
DAQAKD\cite{kur2025data}  &2.41&58.58&69.03& 74.87&17.70\\
CMTKD\cite{pham2023collaborative} &8.96& 63.35&69.58& 72.56&27.10\\
InstantNet\cite{fu2021instantnet} & -- & \textbf{63.60} & 65.61 & 65.44&25.79\\
\midrule
\rowcolor{gray!20}
\textbf{PMABD (Ours)} & \textbf{27.60} & 62.65 & 68.78& 74.63 & 40.50\\
\bottomrule
\vspace{-2.5em}
\end{tabular}
}
\end{table}


\vspace{-0.3cm}
\subsection{Hyperparameter Study Results}
\label{sec:hpsweep_results}
\vspace{-0.2cm}
We also evaluate the performance of our framework on various settings of hyperparameters, as described in Table~\ref{tab:hp_analysis}. It shows that the model improves at a moderate learning rate (lr=$1\times10^{-4}$ ) and large warmup steps ($15$). Distillation temperature is also a critical variable to describe the performance of PMABD.    Table~\ref{tab:hp1} demonstrates the result on $\lambda_{\mathrm{feat}}$ and $\alpha$. The highest accuracy on RestNet-32/CIFAR-100 is achieved at $\lambda_{\mathrm{feat}}=0.3$ and $\alpha=0.65$.
\begin{table}[t]
\centering
\caption{Hyperparameter analysis on ResNet32 / CIFAR-100 (W4A4).
For the learning-rate/warmup study, $\lambda_{\rm feat}=0.3$, $\alpha{=}0.65$, and $\tau=5$.
For the temperature study, $\lambda_{\rm feat}=0.3$, $\alpha{=}0.65$, lr$=5\times10^{-5}$, and warmup$=10$.}
\label{tab:hp_analysis}
\scriptsize
\begin{tabular}{ccc|cc}
\hline
\multicolumn{3}{c|}{\textbf{Learning Rate and Warmup}} &
\multicolumn{2}{c}{\textbf{Temperature}} \\
\hline
\textbf{LR} &
\textbf{Warmup} &
\textbf{Test Acc. (\%)} &
$\boldsymbol{\tau}$ &
\textbf{Test Acc. (\%)} \\
\hline

$1\times10^{-4}$ & 5  & 71.43 &
3 & 70.94 \\

$1\times10^{-4}$ & 15 & \textbf{71.75} & 4 & 70.44 \\

$5\times10^{-5}$ & 5  & 71.15 &
5 & \textbf{71.23} \\

$5\times10^{-5}$ & 15 & 70.69 &
6 & 71.00 \\

$2\times10^{-5}$ & 5  & 70.35 & 7 & 70.71 \\

 $2\times10^{-5}$ & 15 & 70.59 &
8 & 70.44 \\




\hline
\vspace{-3em}
\end{tabular}
\end{table}

\begin{table}[tb]
\vspace{-1em}
\centering
\caption{Effect of KD weight $\alpha$ and  $\lambda_{\rm feat}$
         (ResNet32/CIFAR-100, W4A4). Fixed: $\tau{=}5$, lr$=5{\times}10^{-5}$,
         warmup$=10$.}
\label{tab:hp1}
\resizebox{.28\textwidth}{!}{
\begin{tabular}{c|ccc}
\hline
$\lambda_{\rm feat}$ & $\alpha=0.55$ & $\alpha=0.65$ & $\alpha=0.75$ \\
\hline
0.1 & 70.97 & 70.74 & 71.00 \\
0.3 & 70.87 & \textbf{71.23} & 70.94 \\
0.5 & 70.85 & 71.05 & 71.00 \\
\hline
\end{tabular}
}
\end{table}





\begin{figure}[t]
    \centering
    \includegraphics[width=0.85\linewidth]{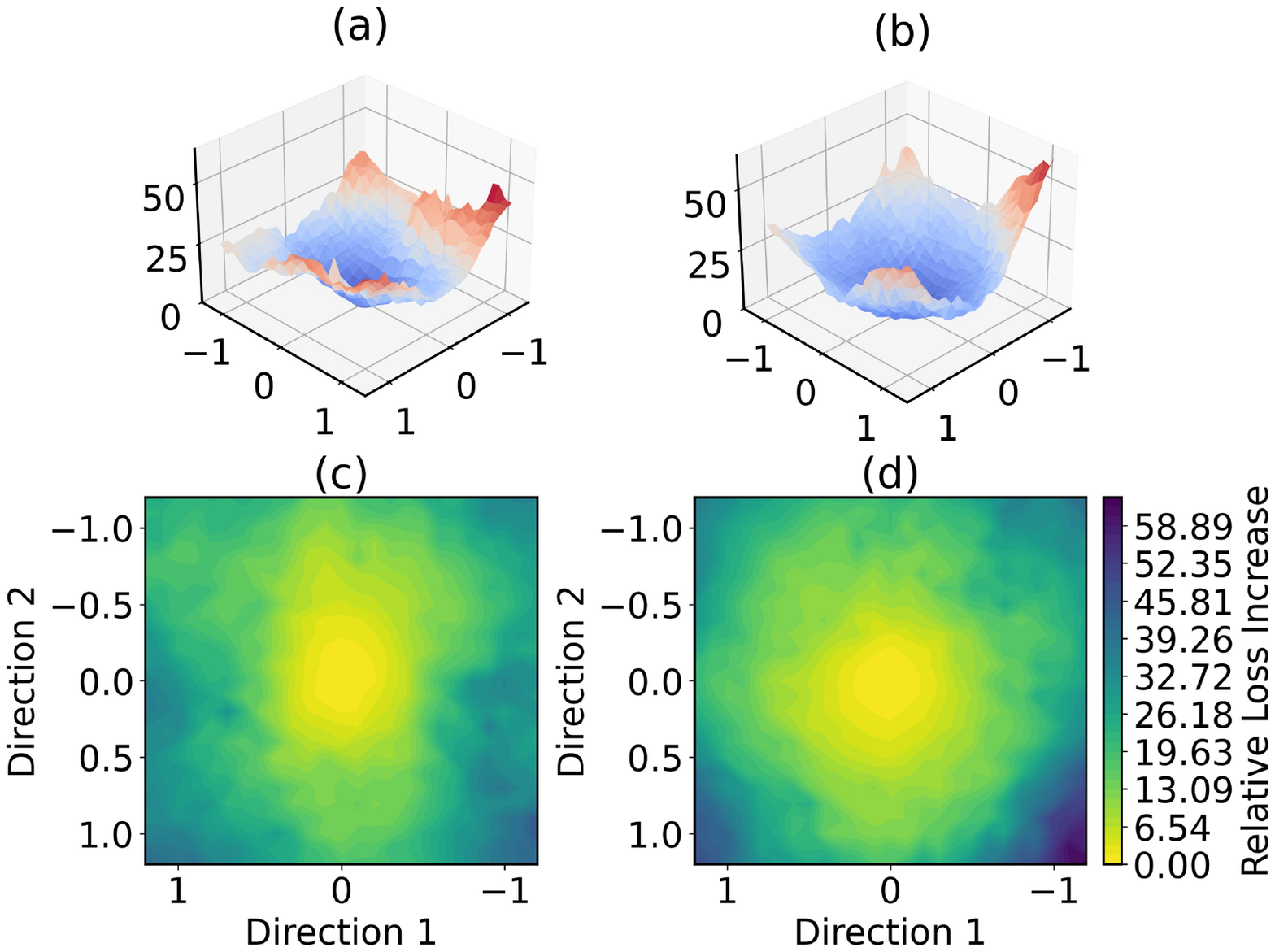}
    \caption{Filter-normalized loss landscape comparison. (a) and (c): Direct 
    W4A4 distillation yields a sharp, irregular, and highly convex minimum. 
    (b) and (d): PMABD W4A4 distillation produces a demonstrably flatter, 
    smoother, and wider basin, explaining its robustness to quantization noise.}
    \label{fig:loss_landscape}
\end{figure}

To validate our core hypothesis that PMABD smoothes the optimization landscape, 
we visualized loss surfaces using filter-normalized weight perturbations 
(Fig.~\ref{fig:loss_landscape}). Direct FP32-to-W4A4 distillation produces a 
sharp, irregular minimum where minor weight perturbations cause steep loss 
penalties. The PMABD W4A4 model occupies a substantially wider and flatter 
basin: the progressive transition through intermediate precisions combined with 
the multi-ancestor supervisory signal acts as a geometric regularizer, keeping 
weights within the convergence zone.
\vspace{-1em}
\subsection{Ablation on Intermediate Teachers}
\vspace{-0.2cm}
To check that every intermediate precision contributes, we train the ResNet-18/CIFAR-100 W2A2 student with progressively larger teacher pools. Each run warm-starts from the lowest-precision model in its pool. As Table~\ref{tab:ablation_pool} shows, test accuracy rises with each teacher added, and the full pool beats distillation from the FP32 teacher alone by 1.33 points. No single stage accounts for the gain; each intermediate model adds to it.

\begin{table}[!tb]
\vspace{-1.2em}
\centering
\caption{Effect of the teacher pool on the W2A2 student (ResNet-18 / CIFAR-100). $\Delta$ is the test accuracy gap relative to our max. accuracy for W2A2 (77.70).}
\label{tab:ablation_pool}
\resizebox{0.9\columnwidth}{!}{
\begin{tabular}{llcc}
\toprule
\textbf{Teacher pool} & \textbf{Warm start}  & \textbf{Test} & \textbf{$\Delta$} \\
\midrule
$\mathcal{M}_1$ (FP32)                                       & $\mathcal{M}_2$ (FP32)   & 76.37 & $-1.33$ \\
$\mathcal{M}_1 + \mathcal{M}_3$ (W8A8)                       & $\mathcal{M}_3$ (W8A8)   & 76.90 & $-0.80$ \\
$\mathcal{M}_1 + \mathcal{M}_3 + \mathcal{M}_4$ (W4A4)       & $\mathcal{M}_4$ (W4A4)   & 77.27 & $-0.43$ \\
$\mathcal{M}_1 + \mathcal{M}_3 + \mathcal{M}_4 + \mathcal{M}_5 + \mathcal{M}_{6a}$  & $\mathcal{M}_{6a}$ (W3A2)  & \textbf{77.70} & -- \\
\bottomrule
\end{tabular}
}
\vspace{-1em}
\end{table}

\section{Conclusion}
\vspace{-0.3cm}
We presented \textbf{Progressive Multi-Ancestor Bit-depth Distillation (PMABD)}, a unified framework that jointly addresses network quantization and knowledge distillation through progressive multi-stage compression. By constructing a sequence of intermediate teachers and enabling each stage to learn from all higher-precision ancestors, PMABD effectively reduces the knowledge gap between teacher and student, leading to more stable optimization during ultra-low-bit quantization. Evaluations on CIFAR-10/100 and Tiny-Imagenet dataset across multiple architectures demonstrate that PMABD achieves superior accuracy while significantly reducing model complexity and memory footprint, outperforming existing SOTA methods. These results suggest that progressive multi-ancestor knowledge transfer is a promising direction for deploying deep neural networks on resource-constrained edge devices.

\let\oldbibliography\thebibliography
\renewcommand{\thebibliography}[1]{%
    \oldbibliography{#1}%
    \setlength{\itemsep}{-1pt}%
    \setlength{\parskip}{0pt}%
}

\bibliographystyle{IEEEbib}
\bibliography{strings,refs}

\end{document}